\documentclass{article} 
\usepackage{iclr2027_conference,times}
\usepackage[T1]{fontenc} 

\usepackage{amsmath,amsfonts,bm}

\def\eqref#1{equation~\ref{#1}}

\def\1{\bm{1}}

\DeclareMathAlphabet{\mathsfit}{\encodingdefault}{\sfdefault}{m}{sl}
\SetMathAlphabet{\mathsfit}{bold}{\encodingdefault}{\sfdefault}{bx}{n}

\usepackage{hyperref}
\usepackage{url}
\usepackage{graphicx}
\usepackage{booktabs}
\usepackage{tabularx}
\usepackage{multirow}
\usepackage{amsthm}
\usepackage{tikz}
\usepackage{algorithm}
\usepackage{algpseudocode}
\usepackage{placeins} 
\usetikzlibrary{arrows.meta,positioning,shapes.geometric}

\newcommand{\enc}{E}
\newcommand{\pred}{F}
\newcommand{\anch}{\mathcal{N}}
\newcommand{\dist}{\lVert q-g\rVert_2}

\title{Latent World Models Should Plan with\\ Distances Along the Latent Manifold}
\hypersetup{hidelinks} 

\author{Zunzhi You$^{1}$, Linwei Tao$^{1,2}$, Siyu Xu$^{1}$, Yunke Wang$^{1}$, Chang Xu$^{1}$\\
$^{1}$The University of Sydney\quad $^{2}$Machine Learning Research, Apple\\
\texttt{zyou5834@uni.sydney.edu.au}\\
\texttt{\{linwei.tao,s.xu,yunke.wang,c.xu\}@sydney.edu.au}}

\iclrfinalcopy
\begin{document}

\maketitle
\lhead{Preprint}

\begin{abstract}
Latent world models such as LeWM plan by searching for an action sequence whose predicted terminal latent is closest to the goal latent under an ambient norm, such as the Euclidean distance. Planning with such a cost assumes that proximity in the ambient latent space reflects proximity to the goal in the environment. Studying LeWM, we find that this assumption holds only in a small neighborhood around the goal, whereas farther away the L2 cost forms spurious low-cost basins that trap the planner. Yet neighborhood-based embeddings show that the latents retain the underlying physical state. We therefore hypothesize that the latents lie near a curved, low-dimensional manifold, and that the terminal cost should measure the distance along this latent manifold rather than through the ambient space. To estimate this manifold distance, we propose the Neighborhood Graph Cost (NGC), which computes shortest-path distances on a mutual nearest-neighbor graph constructed over training latents. NGC changes only the terminal cost and requires no retraining. This single change improves the planning success rate of LeWM across all four benchmark tasks and nearly solves TwoRoom and Reacher: even at the farthest goal offsets tested, NGC reaches $95$\% and $100$\% success, whereas the L2 cost falls to $27$\% and $81$\%. These results support the manifold hypothesis and our argument that latent world models should plan with distances along the latent manifold.
\end{abstract}

\section{Introduction}
\label{sec:intro}

In the paradigm of planning with world models, an agent predicts the outcomes of candidate action sequences and executes the sequence whose predicted outcome scores best \citep{finn2017foresight,hafner2019planet,lecun2022path,hansen2024tdmpc2}.
When the task is specified by a goal observation, the score is naturally a distance between the predicted outcome and the goal, without requiring a separately designed reward.
Because pixel-wise distances to a goal image are dominated by large objects such as the robot arm \citep{ebert2018visualforesight}, latent world models measure the distance in a learned representation space instead \citep{zhou2024dinowm,sobal2025pldm,assran2025vjepa2,terver2026jepawms,maes2026lewm}.
These models typically score the predicted terminal latent against the goal latent using a norm of their difference, such as the Euclidean distance \citep{maes2026lewm} or the L1 distance \citep{assran2025vjepa2}, serving as the sole signal guiding the planner toward the goal.
Planning with such a cost therefore assumes that a latent closer to the goal latent in the latent space corresponds to a state closer to the goal in the environment.

\begin{figure}[t]
\centering
\begin{minipage}[t]{0.56\linewidth}
\centering
\includegraphics[width=\linewidth]{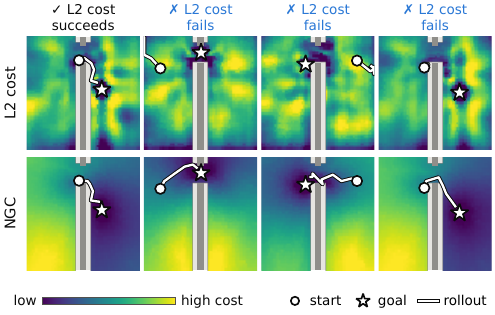}
\caption{
Cost maps and rollout trajectories for four start--goal pairs in TwoRoom.
\textbf{Top:} the L2 cost (Equation~\ref{eq:l2}) between the latent rendered at each feasible grid position and the goal latent.
\textbf{Bottom:} NGC (Section~\ref{sec:method}) on the same pairs. Details in Appendix~\ref{app:costmaps}.}
\label{fig:costmaps}
\end{minipage}\hfill
\begin{minipage}[t]{0.41\linewidth}
\centering
\includegraphics[width=\linewidth]{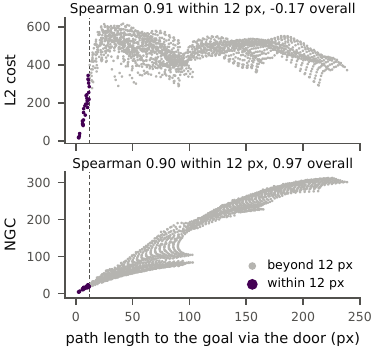}
\caption{
Terminal cost against the length of the shortest feasible path to the goal, for every feasible grid position of the rightmost pair in Figure~\ref{fig:costmaps}.
\textbf{Top:} the L2 cost.
\textbf{Bottom:} NGC.
}
\label{fig:costslice}
\end{minipage}
\end{figure}

\begin{figure}[t]
\centering
\includegraphics[width=\linewidth]{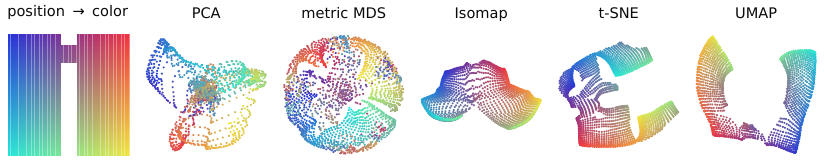}
\caption{Two-dimensional embeddings of the TwoRoom grid latents, with each point colored by its rendered physical position (\textbf{left:} color key).}
\label{fig:projections}
\end{figure}

In this work, we study LeWM \citep{maes2026lewm}, an end-to-end latent world model framework that uses the squared Euclidean distance as its terminal cost (the L2 cost), and find that this assumption fails even on a simple task.
In TwoRoom, a point agent moves from one room to the other through a doorway, and the observation is determined solely by the agent's 2D position. Yet LeWM reaches a success rate of only about $90\%$ when the goal observation is $25$ steps from the start, dropping below $30\%$ at $75$ steps. In every failed episode, the agent becomes trapped in a region far from the goal. The cost maps in Figure~\ref{fig:costmaps} (top) reveal why: while the L2 cost forms a true basin around the goal, it also creates spurious low-cost basins far away that trap the planner.

To understand these basins, we compare the L2 cost against the physical distance to the goal and find that it tracks this distance only locally, ceasing to increase with it farther away (Figure~\ref{fig:costslice}).
Yet neighborhood-based embeddings such as Isomap \citep{tenenbaum2000isomap} unfold the latents into a faithful two-dimensional map of the agent's position, whereas Euclidean-distance-based embeddings fold the space onto itself (Figure~\ref{fig:projections}).
The planning failures, therefore, stem not from the latents losing task-relevant state information, but from how they are read.

These observations lead us to hypothesize that the latents of a trained LeWM lie near a curved, low-dimensional manifold, and that the terminal cost should measure the distance along this latent manifold rather than through the ambient latent space.
To estimate this manifold distance, we propose the \textbf{Neighborhood Graph Cost} (NGC).
NGC connects the latents of training observations to their mutual nearest neighbors, attaches the goal latent and each predicted terminal latent to the graph, and scores a candidate by the shortest-path length between the two. On the TwoRoom pairs, NGC produces a single clean basin at the goal (Figure~\ref{fig:costmaps}, bottom).

We compare NGC with the L2 cost across all four LeWM tasks.
NGC achieves a success rate of $94.6$\% or higher on TwoRoom and $99.4$\% on Reacher, whereas the L2 cost drops to $27$\% and $81$\% at the farthest goal offsets. On Cube and PushT, NGC improves the success rate at the farthest offsets by $16$ and $15$ percentage points.
This gain comes from the cost formulation itself: NGC still outperforms the L2 cost when paired with an oracle predictor that executes every candidate in the simulator, confirming that it does not merely compensate for errors of the learned predictor. Moreover, the L2 cost fails to match NGC even with four times as many CEM candidates.

In summary, we offer a simple explanation for the planning failures we observe in LeWM. It leads to a method that changes only the terminal cost, and the method's effectiveness in turn supports the explanation. We hope that our work leads the latent world model community to reconsider the default of measuring distances through the ambient latent space, and to plan with distances along the latent manifold instead.

\section{Background and related work}
\label{sec:related}

\paragraph{Latent world models for goal-conditioned planning.}
A latent world model encodes observations into a learned latent space and predicts forward in that space, allowing a planner to search over action sequences without rendering pixels \citep{hafner2019planet,lecun2022path}.
When the task is specified by a goal observation, the planner scores each candidate action sequence by a terminal cost between the predicted terminal latent and the goal latent.
DINO-WM \citep{zhou2024dinowm} plans in frozen DINOv2 features;
PLDM \citep{sobal2025pldm} plans in a latent space trained jointly with the predictor;
V-JEPA~2 \citep{assran2025vjepa2} and \citet{terver2026jepawms} plan with JEPA-family world models.
In all cases, the terminal cost is a norm of the latent difference, such as the Euclidean or L1 distance.

We study LeWM \citep{maes2026lewm,maes2026swm}, which trains an encoder $\enc$ and a predictor $\pred$ end-to-end. The encoder maps an observation $o_t$ to a latent $z_t = \enc(o_t)$, and the predictor maps a short history of latents and actions to the next latent, $\hat{z}_{t+1} = \pred(z_{t-2:t},\, a_{t-2:t})$. By rolling $\pred$ forward $H$ steps from the current latent under a candidate action sequence $(a_t, \dots, a_{t+H-1})$, the planner obtains a predicted terminal latent $\hat{z}_{t+H}$.
The planner uses CEM \citep{rubinstein1999cem,deboer2005cem}: it samples candidate action sequences, ranks them by the terminal cost, and iteratively refines the sampling distribution toward lower-cost candidates. Writing $q = \hat{z}_{t+H}$ for the predicted terminal latent and $g = \enc(o_{\mathrm{goal}})$ for the goal latent, the terminal cost is
\begin{equation}
  C_{\mathrm{L2}}(q,g) = \lVert q - g \rVert_2^2.
  \label{eq:l2}
\end{equation}
Because CEM ranks candidates solely by this cost, planning with the L2 cost assumes that a shorter Euclidean distance to the goal latent reflects progress toward the goal in the environment.

\paragraph{Improving LeWM planning.}
Many follow-ups to LeWM aim to improve its planning, largely motivated by the observation that the Euclidean distance between latents fails to measure progress toward the goal.
Their remedies fall into three families.
The first \emph{retrains the encoder} so that the distance between latents itself measures progress, whether by straightening trajectories~\citep{wang2026straightening}, matching latent to ground-truth state distances~\citep{hu2026scale}, adding reachability or trajectory heads~\citep{li2026rcaux,wang2026dalewm,bai2026tdjepa,huang2026trajlewm}, shaping the latent space with a value function~\citep{destrade2025vgjepa}, or embedding in a hyperbolic space~\citep{liu2026proworld}.
The second \emph{learns a parametric cost} on the frozen model, such as a temporal-separation regressor~\citep{li2026trm,singh2026objective}, a goal-conditioned value~\citep{sommer2026rp1}, or a ground-truth state readout~\citep{pham2026leap}.
The third \emph{changes the target} rather than the cost, planning toward a learned subgoal~\citep{masip2026ffjepa,caselli2026hilewm,cheng2026sage} or replacing the search with a learned actor~\citep{sun2026intact}.
Each adds a learned component that must be trained beyond the original LeWM objective.
We depart from all three by attributing the issue to the latents lying near a curved manifold and proposing a method to estimate the distance along it, without training a new encoder, a cost network, a subgoal generator, or a policy.

\paragraph{Manifold learning and shortest-path distances.}
NGC estimates distances along the manifold using ideas from manifold learning.
Isomap \citep{tenenbaum2000isomap} approximates the geodesic distance along a data manifold by the shortest path on a $k$-nearest-neighbor graph weighted by Euclidean edge lengths, then embeds the resulting distance matrix with multidimensional scaling.
t-SNE \citep{vandermaaten2008tsne} and UMAP \citep{mcinnes2018umap} optimize different embedding objectives on similar graphs.
NGC builds the same kind of graph over encoder outputs, but it never computes embeddings or coordinates.
A separate line of work builds graphs over experience and plans by searching them for waypoints \citep{savinov2018sptm,eysenbach2019sorb,yang2020plan2vec,zhang2021l3p}. There, the edges encode demonstrated transitions or learned reachability, and the output is a plan. In contrast, NGC's edges encode only Euclidean proximity, and they are used to score candidates rather than to generate action sequences.

\section{Why LeWM does not solve TwoRoom}
\label{sec:observation}

We begin by examining how LeWM performs on TwoRoom.
Of the four tasks in LeWM, TwoRoom appears to be the easiest: a point agent moves from one room to the other through a fixed doorway, so the top-down observation is determined by the agent's 2D position alone.
Yet under our evaluation protocol (Section~\ref{sec:setup}), LeWM reaches a success rate of only about $90$\% on TwoRoom, compared with $100$\% on Reacher and $91$\% on PushT.
Because the state space is two-dimensional, these failures are also easy to visualize and diagnose.

\paragraph{Spurious low-cost basins trap the agent.}
Visualizing the rollout trajectories reveals that in every failed episode, the agent becomes trapped in a local region for the remainder of the rollout.
To see why the planner chooses such actions, we compute a cost map for each evaluated goal: we place the agent at every feasible position on a $4$\,px grid, render the observation, encode it with the LeWM encoder, and compute its L2 terminal cost against the goal latent (Equation~\ref{eq:l2}).
The top row of Figure~\ref{fig:costmaps} shows four examples overlaid with the executed rollouts.
The cost always forms a basin around the goal, as expected: the cost is zero at the goal, and nearby observations yield similar latents.
Only a short distance away, however, the cost rises sharply, while some regions far from the goal have a much lower cost than their surroundings.
Consequently, when actions sampled by the CEM planner lead into one of these spurious basins rather than the true basin around the goal, the agent becomes trapped.
\begin{figure}[t]
\centering
\includegraphics[width=\linewidth]{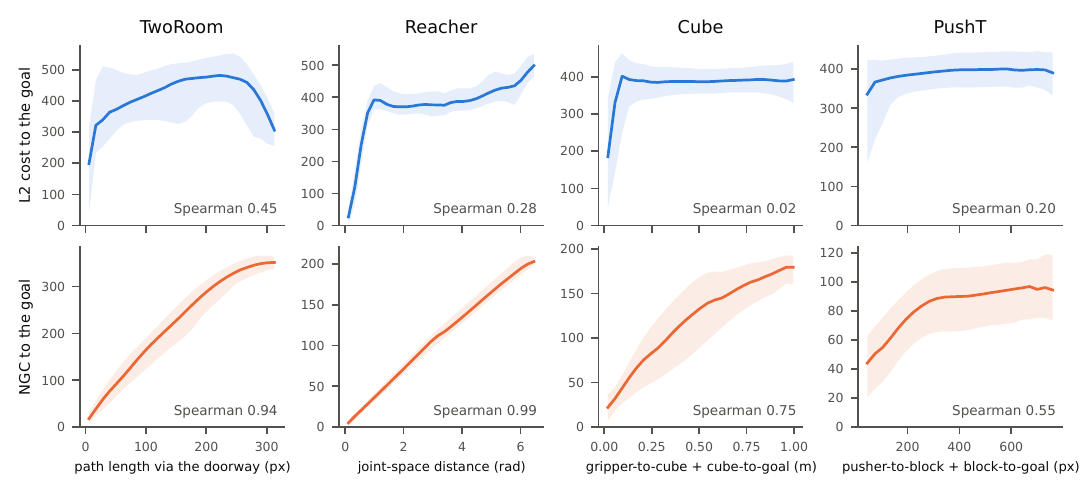}
\caption{Terminal cost against the physical distance to the goal across all four tasks. \textbf{Top:} the L2 cost. \textbf{Bottom:} NGC. Lines show bin medians, and shaded bands indicate the 10th to 90th percentiles. Details in Appendix~\ref{app:costtruth}.}
\label{fig:costtruth}
\end{figure}

\paragraph{The L2 cost is faithful only locally.}
The simplicity of TwoRoom also provides a ``ground-truth'' cost for every position: the length of the shortest feasible path to the goal.
Figure~\ref{fig:costslice} plots the terminal cost against this path length for the rightmost pair in Figure~\ref{fig:costmaps}.
For positions within $12$\,px of the goal along the shortest path, the L2 cost orders positions faithfully (Spearman $0.91$), consistent with the local basin around the goal in Figure~\ref{fig:costmaps}.
Beyond this radius, however, the cost varies widely and no longer increases with path length, dropping the overall Spearman correlation to $-0.17$.

Because the other three tasks have higher-dimensional state spaces, we instead measure how far a state is from the goal using a task-specific \emph{physical distance} computed from the simulator states (Table~\ref{tab:success-criteria}; Appendix~\ref{app:physical}). Across all four tasks, the median L2 cost increases with physical distance only over a short range before flattening out (Figure~\ref{fig:costtruth}, top).

\paragraph{The latents retain the agent's position.}
Why does the L2 cost become uninformative away from the goal? Is it because the encoder fails to capture the task-relevant state, or simply because the cost function fails to read it?
Since a TwoRoom observation is determined by the agent's 2D position, latents that retain this information should admit a faithful two-dimensional embedding.
Figure~\ref{fig:projections} applies five dimensionality reduction methods to the grid latents from Figure~\ref{fig:costmaps}.
Isomap \citep{tenenbaum2000isomap}, t-SNE \citep{vandermaaten2008tsne}, and UMAP \citep{mcinnes2018umap} construct embeddings from each latent's nearest neighbors, and all three unfold the two rooms and recover the physical layout up to a smooth deformation (Spearman $0.81$, $0.83$, and $0.92$ between embedding distance and physical $(x,y)$ distance).
In contrast, principal component analysis (PCA; \citealp{pearson1901pca,hotelling1933pca}), which projects the latents linearly onto their two directions of largest variance, and metric multidimensional scaling (MDS; \citealp{borg2005mds}), which preserves pairwise Euclidean distances, both fold the room onto itself (Spearman $0.52$ and $0.32$).
Thus, the latents do encode the underlying physical state required by the task, but recovering it requires respecting local neighborhoods rather than global Euclidean distances.

\begin{figure}[t]
\centering
\includegraphics[width=\linewidth]{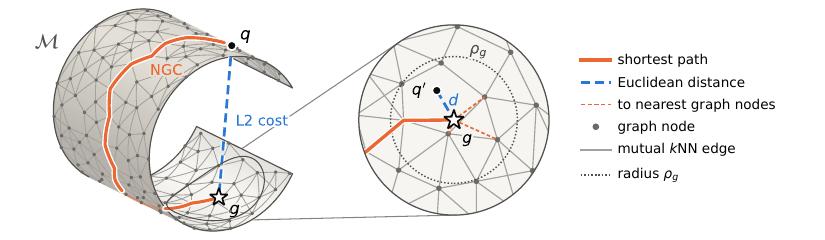}
\caption{NGC on a curved manifold $\mathcal{M}$. \textbf{Left:} a candidate $q$ is close to the goal $g$ in Euclidean distance (blue, the L2 cost) but far from $g$ along $\mathcal{M}$. NGC scores $q$ by the shortest path through the graph (orange), whose edges join mutual $15$-nearest neighbors. \textbf{Right:} the neighborhood of $g$, flattened. When a candidate $q'$ and $g$ lie within each other's neighborhood, NGC scores $q'$ by their Euclidean distance $d=\lVert q'-g\rVert_2$. The surface is a schematic; the LeWM latents are 192-dimensional.}
\label{fig:method}
\end{figure}

\section{Hypothesis and a cost along the latent manifold}
\label{sec:method}

The observations in Section~\ref{sec:observation} lead us to the following hypothesis: the latents of a trained LeWM lie near a curved, low-dimensional manifold $\mathcal{M}$, and the terminal cost should measure the distance along $\mathcal{M}$ rather than through the ambient latent space.

This hypothesis offers a geometric interpretation of all four observations. Whereas the L2 cost measures the squared Euclidean distance between two latents in the ambient space (Equation~\ref{eq:l2}), the distance along $\mathcal{M}$ is the length of the shortest path confined to the manifold.
Within a sufficiently small, nearly flat neighborhood, the Euclidean distance closely approximates this path length, explaining why the L2 cost orders nearby states faithfully.
Over longer distances, however, the manifold curves through the ambient space, so moving farther along $\mathcal{M}$ need not increase the Euclidean distance to the goal.
This accounts for why the L2 cost ceases to increase with physical distance in Figure~\ref{fig:costslice} and Figure~\ref{fig:costtruth} (top).
Moreover, where the manifold curves back toward the goal latent, a region far away along $\mathcal{M}$ can have a lower L2 cost than its surroundings, creating the spurious basins observed in Figure~\ref{fig:costmaps} (top).
Finally, because neighborhood-based embeddings rely on local relations, they can unfold the manifold and recover the underlying physical layout in Figure~\ref{fig:projections}.

Because $\mathcal{M}$ has no explicit parameterization, we cannot compute distances along it analytically and instead estimate them from samples.
On a neighborhood graph whose edges connect only nearby points, each edge approximates a short local segment of $\mathcal{M}$, so the shortest-path length between two nodes estimates the distance along $\mathcal{M}$.
We use this estimator to define the \textbf{Neighborhood Graph Cost} (NGC, Figure~\ref{fig:method}).
Offline, we encode a random subset of $N$ training observations as graph nodes $v_1,\dots,v_N$ and connect each pair of mutual $15$-nearest neighbors with an edge weighted by their Euclidean distance $\lVert v_i-v_j\rVert_2$.
We then precompute all-pairs shortest-path lengths $D_{ij}$ on this between nodes $v_i$ and $v_j$.
Online, we attach the predicted terminal latent $q$ of each candidate action sequence and the goal latent $g$ to their four nearest graph nodes, $\anch_4(q)$ and $\anch_4(g)$, and define the terminal cost as
\begin{equation}
C(q,g)=
\begin{cases}
\dist, & \dist\le\min\{\rho_q,\rho_g\},\\[2pt]
\displaystyle\min_{\substack{i\in\anch_4(q)\\ j\in\anch_4(g)}}\bigl[\lVert q-v_i\rVert_2+D_{ij}+\lVert v_j-g\rVert_2\bigr], & \text{otherwise,}
\end{cases}
\label{eq:ngc}
\end{equation}
where $\rho_q$ and $\rho_g$ denote the distances from $q$ and $g$ to their respective 15th-nearest graph nodes.
When $q$ and $g$ lie within each other's local neighborhood, their Euclidean distance already approximates the manifold distance, so NGC uses it directly.
NGC replaces only the terminal cost; the encoder and predictor remain frozen, and the planner is unchanged.
Algorithm~\ref{alg:ngc} in Appendix~\ref{app:graph} summarizes the procedure.
\section{Experiments}
\label{sec:experiments}

We evaluate NGC by comparing the planning success rate of NGC against the L2 cost on the same trained LeWM.
We first show that NGC improves success rates across all four tasks and that this improvement comes from measuring shortest-path distances on the graph.
We then replace the learned predictor with an oracle predictor to disentangle cost errors from predictor errors. Finally, we evaluate the computational overhead of NGC and test whether increasing the CEM search budget allows the L2 cost to close the gap.

\subsection{Experimental setup}
\label{sec:setup}

We use the four tasks and datasets released with LeWM \citep{maes2026lewm,maes2026swm}: \textbf{TwoRoom}, a point agent navigating between two rooms through a doorway; \textbf{Reacher}, the two-link arm from the DeepMind Control Suite \citep{tassa2018dmcontrol}; \textbf{Cube}, the single-cube manipulation task from OGBench \citep{park2025ogbench}; and \textbf{PushT}, pushing a T-shaped block to a target pose \citep{chi2023diffusionpolicy}. Following the official LeWM release, we train the world model for 10 epochs and adopt the default CEM planner settings (details in Appendix~\ref{app:model}).

\paragraph{Evaluation protocol.} As in  \citet{maes2026lewm}, each evaluation \emph{pair} consists of a start and a goal observation drawn from the same training episode, where the goal is a fixed \emph{offset} of steps after the start, and the agent must reach the goal within twice the offset in primitive steps. We sample pairs with offsets of $25$, $50$, and $100$ ($25$, $50$, and $75$ on TwoRoom, where episodes have $101$ steps). For each task and offset, we evaluate $300$ fixed pairs divided into three sets of $100$. Each set is drawn with its own pair seed and paired with an independently constructed graph, forming three full replicates of the evaluation. Within each set, every pair is evaluated across three planner seeds. We compare the two costs pair by pair: for each pair, we average the success rate over the three planner seeds and compute the difference between NGC and the L2 cost. We report the mean per-pair difference along with a 95\% bootstrap confidence interval (Appendix~\ref{app:prelim}). Appendix~\ref{app:protocol} provides the full protocol.

\subsection{Latent planning performance}
\label{sec:main}

\begin{figure}[t]
\centering
\includegraphics[width=\linewidth]{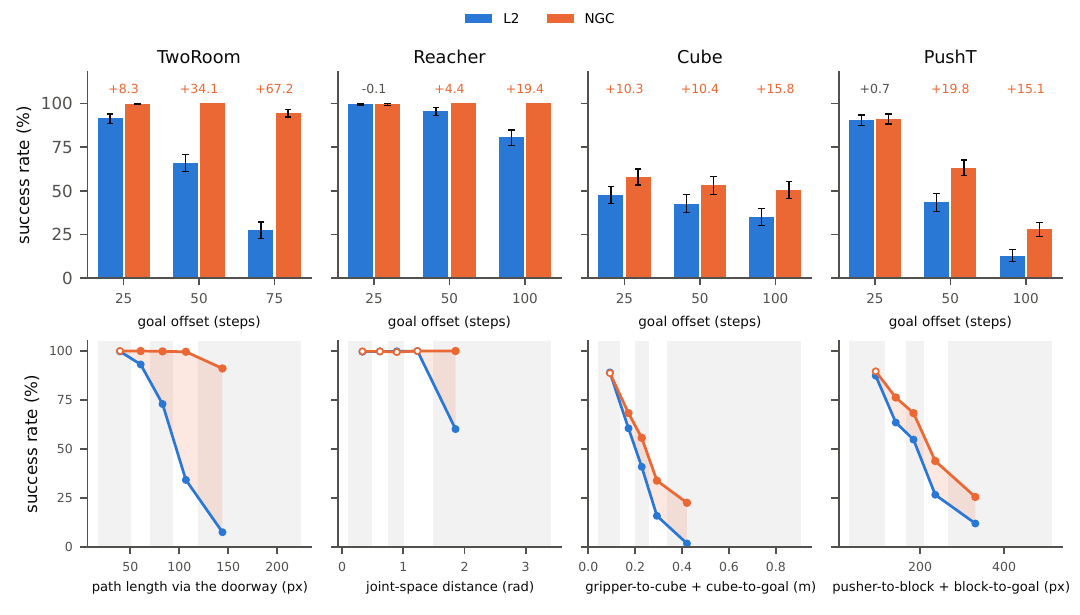}
\caption{\textbf{Top:} Success rate against the goal offset ($300$ pairs per offset; whiskers indicate 95\% bootstrap intervals over pairs). Numbers above the bars show the mean per-pair difference (NGC minus L2, in percentage points), highlighted in orange when the 95\% interval excludes zero. \textbf{Bottom:} The same pairs pooled across offsets and binned into quintiles of the physical distance between the start and goal states ($180$ pairs per bin). Shaded regions show the gap between NGC and the L2 cost, and filled NGC markers indicate bins where the 95\% interval of the difference excludes zero.}
\label{fig:success}
\end{figure}

On TwoRoom and Reacher, the success rate of NGC is at least $99.4$\% at every offset, except $94.6$\% at the farthest on TwoRoom, whereas the L2 cost drops from $91$\% to $27$\% on TwoRoom and from $100$\% to $81$\% on Reacher (Figure~\ref{fig:success}). Replacing the L2 cost with distances along the latent manifold thus nearly solves both tasks. On Cube and PushT, NGC improves the success rate by $10$ to $20$ points at every offset except offset $25$ on PushT (where both costs already exceed $90$\%), though both costs still struggle at the longest offsets. Section~\ref{sec:oracle} examines how much of the remaining failure stems from the learned predictor.

The offset, however, is only a loose proxy for goal distance: the same number of expert steps can move the agent far or barely at all, mixing near and distant goals within each offset. To separate easy and hard goals more directly, the bottom row of Figure~\ref{fig:success} regroups the evaluated pairs into quintiles of the physical distance between the start and goal states. From the nearest quintile to the farthest, the success rate of the L2 cost falls from $99.8$\% to $7.6$\% on TwoRoom and from $99.6$\% to $60$\% on Reacher. In contrast, NGC remains at or above $99.4$\% in every quintile on both tasks, except the farthest quintile on TwoRoom ($91$\%), where the remaining failures arise from predictor errors (Section~\ref{sec:oracle}). On Cube and PushT, the gain is near zero in the nearest quintile and peaks in the farthest two, reaching $21$ and $17$ points. Across all tasks, the L2 cost degrades on physically distant goals, where NGC delivers its largest gains.

Revisiting the diagnostics of Section~\ref{sec:observation} reveals how NGC reshapes the cost landscape that the planner follows. Under NGC, the cost map of each TwoRoom pair forms a single clean basin around the goal (Figure~\ref{fig:costmaps}, bottom), and the cost increases with path length across the entire room (Figure~\ref{fig:costslice}, bottom). Across all four tasks, NGC likewise increases monotonically with physical distance over its full range (Figure~\ref{fig:costtruth}, bottom). Section~\ref{sec:ablation} tests whether the planning gains indeed come from this graph-based distance estimate.

\subsection{Ablation}
\label{sec:ablation}

Besides routing paths along the graph, NGC anchors the cost at graph nodes: the term $\lVert q-v_i\rVert_2$ in Equation~\ref{eq:ngc} is at least the distance from $q$ to its nearest graph node, growing as $q$ moves away from the graph. This raises an alternative explanation for the gain: CEM might exploit predictor errors by selecting candidates whose predicted terminal latent $q$ is close to $g$ in Euclidean space but far from the training latents. Any cost that penalizes such out-of-distribution latents would prevent this exploitation, regardless of whether it measures distance along $\mathcal{M}$. To test this explanation, we evaluate \emph{NGC with Euclidean node distances} (NGC$_{\text{Euc}}$), which replaces $D_{ij}$ in Equation~\ref{eq:ngc} with the direct Euclidean distance $\lVert v_i-v_j\rVert_2$ between the same graph nodes while keeping the rest of NGC unchanged. NGC$_{\text{Euc}}$ applies the exact same anchor penalty to candidates far from the graph.

As shown in Table~\ref{tab:ablation}, NGC$_{\text{Euc}}$ improves over the L2 cost only at offsets $50$ and $75$ on TwoRoom and offsets $50$ and $100$ on Reacher (by $1$ to $14$ points), while elsewhere remaining within $2$ points of the L2 cost or falling $2$ to $5$ points below it. In contrast, replacing $\lVert v_i-v_j\rVert_2$ with shortest-path distances $D_{ij}$ yields substantial gains across all tasks, with 95\% confidence intervals excluding zero everywhere except offset $25$ on Reacher (where the L2 cost already reaches $99.6$\%).  Thus, the improvement comes from measuring distance along the graph, which estimates the manifold distance, rather than from anchoring predicted latents near the training data.

\begin{table}[t]
\centering
\caption{Ablation of the graph shortest-path distance on the pairs and graphs of Figure~\ref{fig:success}. NGC$_{\text{Euc}}$ replaces $D_{ij}$ with the Euclidean distance between the same graph nodes. The last two columns report mean per-pair differences (in percentage points) with 95\% bootstrap intervals.}
\label{tab:ablation}
\footnotesize
\setlength{\tabcolsep}{3pt}
\begin{tabular}{llrrrcc}
\toprule
Task & offset & L2 & NGC$_{\text{Euc}}$ & NGC & NGC$_{\text{Euc}}$ $-$ L2 & NGC $-$ NGC$_{\text{Euc}}$ \\
\midrule
TwoRoom & 25 & 91.4 & 86.6 & 99.8 & $-4.9$ {\scriptsize$[-7.4,-2.3]$} & $+13.2$ {\scriptsize$[+10.0,+16.6]$} \\
 & 50 & 65.9 & 69.0 & 100.0 & $+3.1$ {\scriptsize$[-1.4,+7.8]$} & $+31.0$ {\scriptsize$[+26.2,+36.0]$} \\
 & 75 & 27.3 & 41.3 & 94.6 & $+14.0$ {\scriptsize$[+9.6,+18.6]$} & $+53.2$ {\scriptsize$[+48.0,+58.3]$} \\
\midrule
Reacher & 25 & 99.6 & 99.3 & 99.4 & $-0.2$ {\scriptsize$[-0.9,+0.3]$} & $+0.1$ {\scriptsize$[-0.6,+0.8]$} \\
 & 50 & 95.6 & 96.6 & 100.0 & $+1.0$ {\scriptsize$[+0.2,+2.0]$} & $+3.4$ {\scriptsize$[+1.7,+5.4]$} \\
 & 100 & 80.6 & 82.2 & 100.0 & $+1.7$ {\scriptsize$[+0.2,+3.2]$} & $+17.8$ {\scriptsize$[+13.8,+22.0]$} \\
\midrule
Cube & 25 & 47.6 & 44.1 & 57.9 & $-3.4$ {\scriptsize$[-6.0,-0.9]$} & $+13.8$ {\scriptsize$[+9.9,+17.7]$} \\
 & 50 & 42.7 & 40.9 & 53.1 & $-1.8$ {\scriptsize$[-4.3,+0.7]$} & $+12.2$ {\scriptsize$[+8.4,+16.2]$} \\
 & 100 & 34.8 & 33.0 & 50.6 & $-1.8$ {\scriptsize$[-4.6,+1.0]$} & $+17.6$ {\scriptsize$[+13.2,+22.0]$} \\
\midrule
PushT & 25 & 90.6 & 88.2 & 91.2 & $-2.3$ {\scriptsize$[-4.8,0.0]$} & $+3.0$ {\scriptsize$[+1.3,+4.9]$} \\
 & 50 & 43.3 & 40.7 & 63.1 & $-2.7$ {\scriptsize$[-6.8,+1.4]$} & $+22.4$ {\scriptsize$[+17.8,+27.2]$} \\
 & 100 & 12.8 & 9.8 & 27.9 & $-3.0$ {\scriptsize$[-5.3,-0.8]$} & $+18.1$ {\scriptsize$[+14.8,+21.7]$} \\
\bottomrule
\end{tabular}
\end{table}

\begin{table}[t]
\centering
\caption{Success rate (\%) with the learned predictor and with the oracle predictor, which executes each candidate in the simulator and encodes its terminal frame, on one set of $100$ pairs with a single planner seed. The last two columns report mean per-pair differences in \% with 95\% bootstrap intervals: NGC minus L2 under the oracle predictor, and oracle minus learned predictor under NGC.}
\label{tab:oracle}
\small
\begin{tabular}{llcccccc}
\toprule
& & \multicolumn{2}{c}{L2} & \multicolumn{2}{c}{NGC} & oracle & NGC \\
\cmidrule(lr){3-4}\cmidrule(lr){5-6}
Task & offset & learned & oracle & learned & oracle & NGC $-$ L2 & oracle $-$ learned \\
\midrule
TwoRoom & 25 & 92 & 84 & 100 & 100 & $+16$ {\scriptsize$[+9,+23]$} & $0$ \\
TwoRoom & 50 & 65 & 58 & 100 & 100 & $+42$ {\scriptsize$[+32,+52]$} & $0$ \\
TwoRoom & 75 & 24 & 19 & 94 & 100 & $+81$ {\scriptsize$[+73,+88]$} & $+6$ {\scriptsize$[+2,+11]$} \\
Cube & 25 & 44 & 56 & 59 & 76 & $+20$ {\scriptsize$[+11,+29]$} & $+17$ {\scriptsize$[+8,+26]$} \\
Cube & 50 & 44 & 52 & 53 & 71 & $+19$ {\scriptsize$[+11,+27]$} & $+18$ {\scriptsize$[+11,+26]$} \\
Cube & 100 & 31 & 38 & 50 & 74 & $+36$ {\scriptsize$[+26,+46]$} & $+24$ {\scriptsize$[+15,+33]$} \\
PushT & 25 & 90 & 94 & 88 & 95 & $+1$ {\scriptsize$[0,+3]$} & $+7$ {\scriptsize$[+2,+13]$} \\
PushT & 50 & 43 & 58 & 61 & 84 & $+26$ {\scriptsize$[+16,+36]$} & $+23$ {\scriptsize$[+14,+33]$} \\
PushT & 100 & 14 & 13 & 28 & 36 & $+23$ {\scriptsize$[+15,+32]$} & $+8$ {\scriptsize$[-2,+18]$} \\
\bottomrule
\end{tabular}
\end{table}

\subsection{Planning with an oracle predictor}
\label{sec:oracle}

All results so far use the learned predictor, whose rollout errors could affect both the gains of NGC and its remaining failures. To isolate the role of the terminal cost, we replace the learned predictor with an oracle predictor: for each CEM candidate, the simulator executes the action sequence from a copy of the exact environment state, and the frozen encoder encodes the resulting terminal frame. The encoder, graph, cost, and planner remain unchanged. Because querying the simulator for all $300$ candidates across $30$ CEM iterations is computationally expensive, we evaluate the oracle on one set of $100$ pairs with a single planner seed, omitting Reacher where NGC already reaches $100$\% at the two longer offsets. Appendix~\ref{app:oracle} provides further details.

If NGC merely compensated for errors of the learned predictor, the two costs would achieve similar success rates under the oracle. Instead, NGC still outperforms the L2 cost by $16$ to $81$ points in eight of the nine settings in Table~\ref{tab:oracle}. The gain of NGC therefore comes from correcting the terminal cost itself, not from compensating for an imperfect predictor. Notably, on TwoRoom, replacing the learned predictor with the oracle actually lowers the success rate of the L2 cost by $5$ to $8$ points (analyzed in Appendix~\ref{app:oraclel2}).

The oracle predictor also shows how much of NGC's remaining failure stems from prediction errors, since such failures disappear once predictions are exact. On TwoRoom, prediction errors account for  all remaining failures: paired with the oracle, NGC solves $100$\% of the pairs across all offsets, including the $6$ pairs at offset $75$ where rollouts with the learned predictor get stuck against the wall beside the doorway. On Cube and PushT, the oracle predictor raises the success rate of NGC by $7$ to $24$ points, showing that predictor errors also account for a substantial portion of the failures on these tasks. Even with exact predictions, however, NGC remains far from solving Cube and PushT, pointing to limitations beyond the predictor (Section~\ref{sec:discussion}).

\subsection{Computation and planning time}
\label{sec:efficiency}

\begin{table}[t]
\centering
\caption{Planning time of NGC at offset $25$. CEM: median time of one CEM optimization, with its ratio to L2. Evaluation: total wall-clock time of a full evaluation relative to L2 on the same pairs. Objective: median time to score one pool of $300$ candidates (Appendix~\ref{app:efficiency}).}
\label{tab:efficiency}
\small
\setlength{\tabcolsep}{4pt}
\begin{tabular}{lcccrr}
\toprule
& \multicolumn{2}{c}{CEM (s)} & Evaluation & \multicolumn{2}{c}{Objective (ms per pool)} \\
\cmidrule(lr){2-3}\cmidrule(lr){4-4}\cmidrule(lr){5-6}
Task & L2 & NGC & ($\times$ L2) & L2 & NGC \\
\midrule
TwoRoom & 0.49 & 0.48 {\scriptsize($0.98\times$)} & 0.86 & 0.071 & 0.67 \\
Reacher & 0.49 & 0.52 {\scriptsize($1.06\times$)} & 1.05 & 0.070 & 0.66 \\
Cube & 0.49 & 0.53 {\scriptsize($1.07\times$)} & 1.08 & 0.070 & 0.67 \\
PushT & 0.50 & 0.52 {\scriptsize($1.05\times$)} & 1.13 & 0.070 & 0.67 \\
\bottomrule
\end{tabular}
\end{table}

\begin{figure}[t]
\centering
\includegraphics[width=\linewidth]{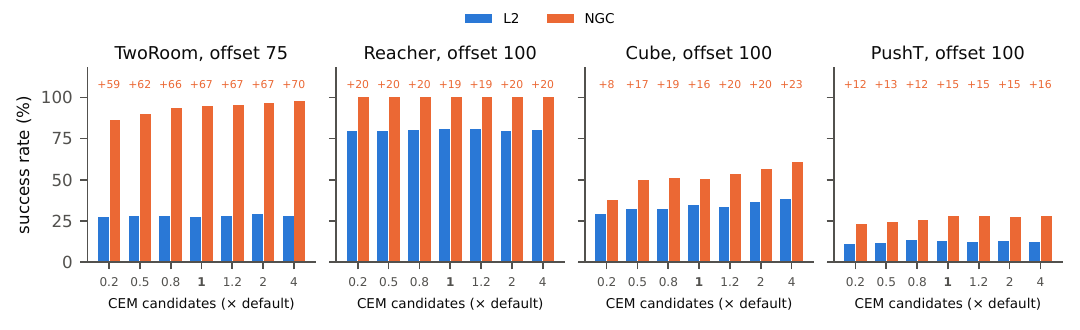}
\caption{Success rate against the number of CEM candidates at the longest offset of each task. The number of elites scales proportionally; $1\times$ corresponds to the default setting in Figure~\ref{fig:success}. The elite fraction and the $30$ CEM iterations remain fixed, and all multipliers are evaluated on the pairs, planner seeds, and graphs of Figure~\ref{fig:success}.}
\label{fig:budget}
\end{figure}

Compared with the L2 cost, NGC introduces additional computation, but nearly all of it takes place offline. The shortest paths between all graph nodes are precomputed once after the model is trained, and attaching the goal latent to the graph is done once per episode. During planning, the online overhead of scoring candidates is small compared to predictor rollouts. Consequently, a single CEM optimization with NGC takes at most $7$\% longer than with the L2 cost, and a full evaluation takes $0.86\times$ to $1.13\times$ as long, running faster on TwoRoom because successful episodes terminate early (Table~\ref{tab:efficiency}).

Conversely, giving the L2 cost more compute does not close the performance gap. Figure~\ref{fig:budget} scales the search budget by varying the number of CEM candidates from $0.2\times$ to $4\times$ the default at the longest offset of each task. Increasing the budget fourfold adds far more computation than NGC's online overhead, yet it improves the success rate of the L2 cost by no more than $3.2$ points, leaving it well below default NGC on every task. What limits planning with the L2 cost is the cost geometry, not the search budget.

\section{Discussion and conclusion}
\label{sec:discussion}

\paragraph{Where does the manifold come from?}
LeWM is trained solely for next-latent prediction and regularization, without any supervision over state-to-state distances. To make latent distances plannable, many follow-ups therefore introduce explicit long-range supervision, such as temporal reachability or privileged physical state (Section~\ref{sec:related}).
Yet without any such supervision, distances along the latent manifold of LeWM suffice to nearly solve TwoRoom and Reacher and yield consistent gains across all four tasks.
Nearby states often produce similar observations and respond similarly to the same actions.
We conjecture that predictive training preserves this local continuity while preventing representation collapse, naturally yielding latent neighborhoods that mirror the local structure of the state space.

\paragraph{Limitations.}
This local continuity hypothesis also points to a natural limitation at contact, where nearby physical states can respond very differently to the same action: a gripper a few millimeters away from a cube cannot lift it, whereas one grasping it can. If the encoder places such states in the same latent neighborhood, changing the terminal cost alone may not resolve the distinction. Indeed, NGC reaches near-perfect success on TwoRoom and Reacher, which involve smooth motion, but still falls short on the contact-rich Cube and PushT tasks even with an oracle predictor. Furthermore, contact introduces asymmetric transitions that an undirected graph cannot express. Finally, our evaluation is confined to the training distribution: following LeWM \citep{maes2026lewm}, we draw start--goal pairs from episodes used to train the world model (though we exclude evaluated episodes from graph construction). Generalization to unseen environments or goals outside the coverage of the graph remains untested.

\paragraph{Future work.}
We plan to investigate how latent neighborhoods can better capture contact dynamics and the directionality of control. Another promising direction is to construct and evaluate NGC using data collected from random or autonomous exploration. Because NGC relies solely on observation latents without requiring actions or trajectory order, neither the graph nor the world model's prediction objective depends on expert demonstrations. Exploration data thus offers a natural next test of whether our findings extend beyond the demonstration datasets used here.

\paragraph{Conclusion.} Many efforts to extend latent world models to longer horizons treat planning failures as a representation failure, retraining the encoder so that Euclidean distances measure task progress.
Our results suggest that much of this failure lies not in what the representation encodes, but in how the planner reads it.
Predictive training alone already yields faithful local neighborhoods, and a shortest-path graph over these neighborhoods turns them into an effective planning cost without any additional training.
We therefore argue that latent world models should be built and evaluated around the geometry of their latent manifold, rather than expecting global Euclidean distances in the ambient space to measure progress directly.

\subsubsection*{AI use statement}
In this work, we used LLM-based assistants to search for and summarize related literature, implement experiments, create visualizations, and translate the authors' intention to academic English writing.
We have reviewed all AI-assisted work. Every number in the paper is copied from archived experiment reports, and every figure is generated by a script from those records. Every citation was checked against the cited work.
We are aware that AI-generated text can be painful to read, and we have tried our best to ensure that the presentation is clear, accurate, and readable to human readers.

\subsubsection*{Ethics statement}
This work uses only simulated benchmark environments and publicly released datasets and checkpoints. It involves no human subjects, personal data, or deployment on physical systems. We see no direct ethical concerns beyond those common to robot-learning research.

\subsubsection*{Reproducibility statement}
All experiments use the public \texttt{stable-worldmodel} package and the released LeWM datasets. Section~\ref{sec:setup} and Appendix~\ref{app:setup} specify the checkpoints, the fixed pair sets, the graph construction rule, the planner budget, the numerical settings, and the statistical procedure. Every reported cell is backed by a per-pair outcome record. The code is available at \url{https://github.com/youzunzhi/ngc-latent-manifold}.

\bibliography{references}
\bibliographystyle{iclr2027_conference}

\clearpage
\appendix
\begin{algorithm}[t]
    \caption{Neighborhood Graph Cost (NGC)}
    \label{alg:ngc}
    \begin{algorithmic}[1]
    \Statex \textbf{Offline}, once per encoder
    \Require a random subset of $N$ training observations $o_1,\dots,o_N$
    \State $v_i\gets\enc(o_i)$ for $i=1,\dots,N$ \Comment{graph nodes}
    \State join $i$ and $j$ by an edge of length $\lVert v_i-v_j\rVert_2$ if each is among the 15 nearest graph nodes of the other
    \While{the graph has more than one component}
    \State $(i,j)\gets\arg\min_{c(i)\neq c(j)}\lVert v_i-v_j\rVert_2$ \Comment{$c(i)$: component of $v_i$}
    \State join $i$ and $j$ by an edge of length $\lVert v_i-v_j\rVert_2$ \Comment{merges $c(i)$ and $c(j)$}
    \EndWhile
    \State $D_{ij}\gets$ shortest-path length between $i$ and $j$ for all $i,j$ \Comment{all-pairs shortest paths via Dijkstra}
    \Statex
    \Statex \textbf{Online}, once per goal
    \Require goal latent $g$
    \State $\anch_4(g)\gets$ the 4 nearest graph nodes of $g$;\quad $\rho_g\gets$ distance from $g$ to its 15th-nearest graph node
    \Statex \textbf{Online}, for each candidate scored by CEM
    \Require predicted terminal latent $q$
    \State $\anch_4(q)\gets$ the 4 nearest graph nodes of $q$;\quad $\rho_q\gets$ distance from $q$ to its 15th-nearest graph node
    \State $d\gets\dist$
    \If{$d\le\min\{\rho_q,\rho_g\}$} \Return $d$ \Comment{Equation~\ref{eq:ngc}, first case}
    \EndIf
    \State \Return $\min_{i\in\anch_4(q),\,j\in\anch_4(g)}\bigl[\lVert q-v_i\rVert_2+D_{ij}+\lVert v_j-g\rVert_2\bigr]$ \Comment{Equation~\ref{eq:ngc}, second case}
    \end{algorithmic}
\end{algorithm}

\section{Implementation and experimental details}
\label{app:setup}

\subsection{World model and planner}
\label{app:model}

\paragraph{World model.} The world models we train follow the released LeWM recipe. The encoder is a ViT-tiny (patch size 14, image size 224, width 192) trained from scratch. The predictor is a six-layer transformer (16 heads, MLP width 2048) conditioned on a history of three frames. The action embedder has hidden width 768, the two projector heads have hidden width 2048, and SIGReg has weight 0.09 (17 knots, 1024 projections). Training uses AdamW (learning rate $5\times10^{-5}$ with a linear warmup over the first 1\% of steps and cosine decay, weight decay $10^{-3}$), batch size 128, bf16 precision, gradient clipping at 1.0, and frameskip 5. Every task is trained for 10 epochs, over which the learning-rate schedule completes. Unless stated otherwise, we use the epoch-10 checkpoint (our checkpoint). We also evaluate the checkpoints released by \citet{maes2026lewm} on HuggingFace (\texttt{quentinll/lewm-*}) to verify that our evaluator reproduces their published success rates (Appendix~\ref{app:comparability}) and to test whether the gains of NGC generalize across checkpoints (Appendix~\ref{app:ckpt}).

\paragraph{Planner.} CEM uses identical settings across all tasks. A plan spans $H=5$ predictor steps of five primitive actions each ($25$ primitive steps). Each of the $30$ iterations samples $300$ candidate action sequences and refits the sampling distribution to the $30$ elites with the lowest terminal cost. The planner executes all five blocks of the final elite mean before running the next CEM optimization from the resulting state. An episode terminates at the first primitive step that meets the success criterion, or after twice the offset in primitive steps.

\subsection{NGC and its graph}
\label{app:graph}

For each task and pair seed, the graph nodes consist of $N=5{,}000$ training observations drawn from distinct episodes that are excluded from evaluation under that seed. Each node is a single random frame from its episode, and every checkpoint encodes the exact same set of observations. If the mutual-15NN graph of a checkpoint has multiple connected components, we iteratively join them via their shortest inter-component edges (Algorithm~\ref{alg:ngc}), leaving all other edges unchanged.

\paragraph{Computing the cost.} Our implementation stores the node latents and the all-pairs shortest-path table $D$ in single precision on the planning device. For $N=5{,}000$, this requires $103.8$\,MB of memory: $N^2\times4$\,B $=100$\,MB for $D$ and $N\times192\times4$\,B $=3.8$\,MB for the latents. Because the goal's nearest graph nodes, their distances to the goal, and $\rho_g$ remain fixed throughout an episode, they are computed once per goal. Scoring a candidate online then requires only a single nearest-neighbor query and taking the minimum over the $16$ anchor pairs in Equation~\ref{eq:ngc}.

\paragraph{Timing.}\label{app:efficiency} Table~\ref{tab:efficiency} benchmarks this computation on a single RTX 4090 GPU. The objective time is measured over $200$ pools of $300$ candidates per graph and averaged across the three graphs of each task. The CEM and evaluation times correspond to the offset-$25$ evaluations in Table~\ref{tab:success}. Because total evaluation time sums across all CEM optimizations in an evaluation, it also reflects how many planning steps episodes require before terminating. NGC runs fewer optimizations than the L2 cost on TwoRoom ($938$ versus $1{,}070$) and Cube ($1{,}341$ versus $1{,}406$) because it solves more episodes early, while running the same number of optimizations on Reacher and PushT.

\subsection{Evaluation protocol}
\label{app:protocol}

\paragraph{Seeds.} Three types of random seeds structure our evaluation. A \emph{pair seed} draws the evaluation pairs for a task (one set of $100$ pairs per offset, yielding $300$ pairs in total across three pair seeds) and samples the nodes of the corresponding NGC graph; the three pair seeds thus form three full replicates of the evaluation. A \emph{planner seed} controls the CEM sampling distribution; every pair is evaluated across three planner seeds, which serve as repeated measurements and are averaged within each pair (Appendix~\ref{app:prelim}). Finally, a \emph{training seed} controls the world model training run. All evaluation pair sets were drawn after NGC and the evaluation protocol were finalized, independently of the pairs used during method development.

\paragraph{Pair sets.} Each task, offset, and pair seed defines a fixed set of $100$ pairs drawn from $100$ distinct episodes. A candidate pair consists of any start step in a dataset episode and the goal observation located one offset later in the same episode. Following \citet{sun2026intact}, we discard candidates whose start state already satisfies the goal's success criterion, and on TwoRoom we additionally discard candidates whose start and goal lie in the same room. We sample $100$ distinct episodes uniformly from those retaining at least one valid candidate pair after filtering, and then sample one valid start step uniformly within each selected episode.

\paragraph{Execution and success.} Each episode initializes the simulator to the exact state of the start observation, except that all velocities are zeroed out. Because the planner observes only the single start image, which conveys pose but not velocity, a nonzero initial velocity would cause unobservable drift that the planner cannot anticipate. The first planning step uses this single-frame context, while subsequent CEM optimizations use the executed history of three frames at stride 5, matching the training configuration. Table~\ref{tab:success-criteria} lists the task-specific success criteria. On Reacher, we compare the shoulder angle on the unit circle as proposed by \citet{sun2026intact}: because the shoulder is an unlimited hinge, angles separated by a full rotation represent the exact same physical configuration and render the same image. Appendix~\ref{app:comparability} also evaluates the unmodified LeWM protocol for direct comparison with published results.

\paragraph{Physical distance.}\label{app:physical} To measure how far a state is from the goal in the environment, we compute a task-specific \emph{physical distance} from simulator states (Table~\ref{tab:success-criteria}). On TwoRoom, this is the shortest feasible path length via the doorway; on Reacher, it is the Euclidean distance in joint space, so both directly measure the remaining motion. On Cube and PushT, the metric approximates the remaining work: it sums the end-effector-to-object and object-to-goal distances, without accounting for grasping dynamics or the detour required to position the pusher behind the block. In the bottom row of Figure~\ref{fig:success}, we bin evaluation pairs by the physical distance between their start and goal states, as a fixed number of expert steps can correspond to either large or negligible physical displacement. We also use this metric in Appendix~\ref{app:costtruth} to evaluate how faithfully each terminal cost orders states.

\begin{table}[h]
\centering
\caption{Success criteria and physical distances (Appendix~\ref{app:physical}). An episode succeeds at the first step that meets the criterion.}
\label{tab:success-criteria}
\small
\begin{tabularx}{\linewidth}{@{}l>{\raggedright\arraybackslash\leavevmode}X>{\raggedright\arraybackslash\leavevmode}X@{}}
\toprule
Task & Success criterion & Physical distance \\
\midrule
TwoRoom & Endpoint within 16\,px of the goal position & Path length via the doorway \\
\addlinespace
Reacher & Joint error $<0.05$\,rad, shoulder angle compared on the circle & Norm of the difference of the two joint angles, shoulder angle compared on the circle \\
\addlinespace
Cube & Cube center within 4\,cm of the goal position & Gripper-to-cube plus cube-to-goal distance \\
\addlinespace
PushT & Pusher and block within 20\,px of the goal pose and block angle within $20^\circ$ & Pusher-to-block distance plus the root-mean-square displacement over the block's area to the goal pose \\
\bottomrule
\end{tabularx}
\end{table}

\subsection{Statistics}
\label{app:prelim}

Each pair is evaluated under both costs using the same three planner seeds. Averaging success over the three seeds yields a per-pair success rate for each cost, and their difference (NGC minus the L2 cost) takes a value in $\{0, \pm\frac13, \pm\frac23, \pm1\}$ (Figure~\ref{fig:pairdiff}). We denote the mean per-pair difference (in \%) by $\Delta$. Its 95\% confidence interval estimates how much $\Delta$ would vary across new draws of evaluation pairs. We compute this interval via a paired bootstrap over the evaluated pairs: a \emph{resample} draws, with replacement, as many pairs as were evaluated in the comparison ($300$ in Table~\ref{tab:success}), and we take the 2.5th and 97.5th percentiles of $\Delta$ across $50{,}000$ such resamples. The interval reflects variation in the per-pair differences rather than in either cost's absolute success rate. Pairs on which both costs perform identically contribute zero, while the others contribute positive or negative differences (the colored segments in Figure~\ref{fig:pairdiff}). The interval width depends on the distribution of these differences, including the proportion of zeros.
The interval reflects the variation across evaluation pairs for the evaluated checkpoint, graphs, and planner seeds; Tables~\ref{tab:seeds}, \ref{tab:graphsens}, and~\ref{tab:graphk} report the variation across training seeds and graph construction. Intervals are not corrected for multiple comparisons. Where every pair has the same value, such as an arm that succeeds on every pair, all resamples coincide and we report no interval. The physical-distance bins of Figure~\ref{fig:success} pool the offsets and can hold several pairs of one episode, so their resamples draw episodes instead of pairs.

\subsection{Oracle predictor}
\label{app:oracle}

The oracle predictor in Section~\ref{sec:oracle} replaces only the learned predictor inside CEM. To score a candidate action sequence, it copies the current simulator state of the ongoing episode into a secondary simulator instance, steps the candidate forward, renders the terminal frame, and encodes it with the frozen encoder. The exact state copied depends on the environment: on TwoRoom, it includes the agent and target positions, wall and doorway parameters, and rendering variation variables; on Cube, it is the full MuJoCo physics data structure. On PushT, directly setting the state does not restore the internal state of the physics solver, so the oracle instead resets the secondary simulator and replays all actions executed in the episode so far. To match the fixed rollout horizon of the learned predictor, the secondary simulator executes each candidate action sequence in full, even if the agent passes through the goal state midway. The scored terminal frame thus always corresponds to the state after the final action of the candidate sequence. We run the oracle evaluation on the pair sets of the first pair seed, using its corresponding graph and a single planner seed; the learned-predictor columns of Table~\ref{tab:oracle} are reported on the exact same pairs and planner seed.


\section{More details of the cost diagnostics}
\label{app:offline}

\subsection{TwoRoom diagnostics}
\label{app:costmaps}

\begin{figure}[t]
\centering
\includegraphics[width=\linewidth]{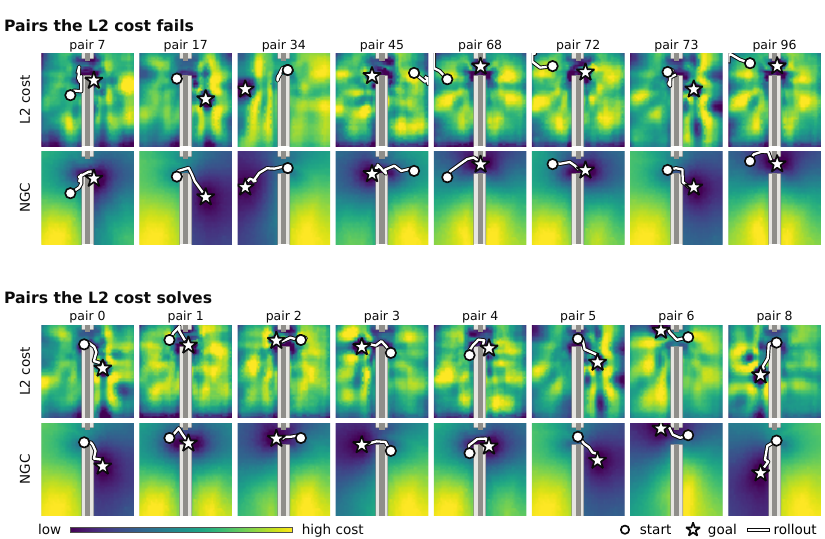}
\caption{Cost maps and rollout trajectories for additional TwoRoom pairs, formatted as in Figure~\ref{fig:costmaps}, with the L2 cost map above the NGC map of each pair. \textbf{Upper block:} all $8$ pairs from the evaluation of Figure~\ref{fig:costmaps} on which the L2 cost fails. \textbf{Lower block:} the first $8$ pairs in pair-index order on which the L2 cost succeeds. NGC solves every pair shown.}
\label{fig:costmapgallery}
\end{figure}

\paragraph{Cost maps.} The cost maps in Figure~\ref{fig:costmaps} evaluate $C(\enc(\mathrm{render}(x,y)),\enc(o_{\mathrm{goal}}))$ using our checkpoint on a $46\times46$ grid with a $4$\,px stride, restricted to the $1{,}882$ collision-free positions where the simulator allows the agent center to be placed. Each image pixel is assigned the cost of its nearest grid point (at most $2$\,px away). Each panel is normalized independently to its own 2nd--98th percentile range, so color scales are comparable only within a single panel. Dark gray indicates the walls, light gray marks valid grid positions that the environment dynamics cannot reach, and the white trajectory curve connects the agent positions recorded after each planner step. Both rows are taken from the offset-$25$ evaluation on the first pair set (the same set used for the oracle experiments), where the L2 cost solves $92$ pairs and NGC solves all $100$. The four columns show one pair where the L2 cost succeeds and three of the $8$ pairs where it fails. In these three failures, the L2 rollout terminates in the cheapest $5$\%, $30$\%, and $3$\% of the cost map, despite remaining $93$, $109$, and $83$\,px away from the goal along the shortest path (trapped in a low-cost band along a far corner of the start room, against its far wall, and in a pocket beside the doorway, respectively). Figure~\ref{fig:costmapgallery} shows cost maps for all $8$ failures of the L2 cost in this evaluation alongside the first $8$ successes in pair-index order.

\paragraph{Projections.} Figure~\ref{fig:projections} shows two-dimensional embeddings of the $1{,}882$ grid latents produced by five dimensionality reduction methods. Each embedding is fit once on the grid latents and plotted directly. The first two methods operate on pairwise Euclidean distances: PCA projects the unwhitened latents onto their top two principal components, while metric MDS minimizes the raw stress $\sum_{i<j}(\lVert y_i-y_j\rVert-\lVert z_i-z_j\rVert)^2$ via SMACOF initialized from the classical MDS solution, reaching a Kruskal stress-1 of $0.375$. The remaining three methods build on local nearest neighbors: Isomap uses an undirected $8$-nearest-neighbor union graph followed by classical MDS; t-SNE uses perplexity 50, PCA initialization, and a fixed random seed; and UMAP uses 15 neighbors, a minimum distance of 0.1, and a fixed random seed.

\subsection{Cost and physical distance}
\label{app:costtruth}

On tasks where the state space cannot be exhaustively gridded as in TwoRoom (Figure~\ref{fig:costslice}), Figure~\ref{fig:costtruth} uses the task-specific physical distance (Appendix~\ref{app:physical}) as a reference to test how well each cost orders states by their distance to the goal. For each task, we sample $10{,}000$ frames uniformly from dataset episodes not used in graph construction and evaluate all $10{,}000\times9{,}999 \approx 10^8$ ordered frame--goal pairs (treating the first frame as the state and the second as the goal). Because physical distances are computed directly from the underlying simulator states, this diagnostic is purely offline and involves no planning. Pairs are grouped into equal-width bins spanning from zero to the maximum observed physical distance; bins containing fewer than $1{,}000$ pairs (accounting for fewer than $700$ pairs in total per task) are omitted. Across the other two pair seeds, the resulting Spearman correlations remain within $0.03$ of those reported in Figure~\ref{fig:costtruth}.


\section{More results on planning performance}
\label{app:planning}

\begin{figure}[h]
\centering
\includegraphics[width=\linewidth]{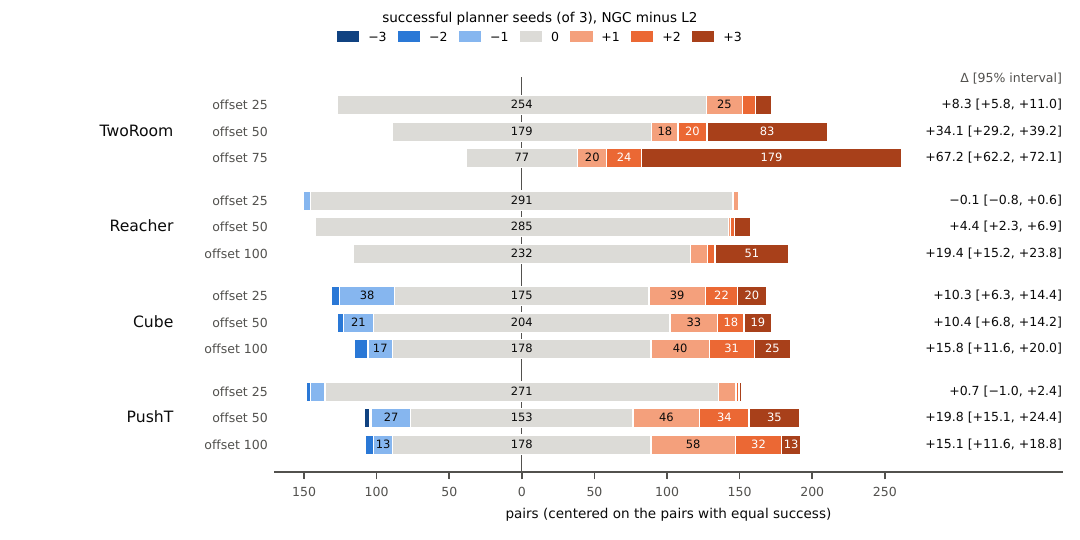}
\caption{Per-pair differences underlying Table~\ref{tab:success}. Each bar represents the $300$ pairs for a given task and offset. Under each cost, every pair is evaluated with three planner seeds; colors indicate the number of seeds on which NGC succeeds minus the number on which the L2 cost succeeds. Gray segments (equal success under both costs) are centered at zero; pairs where NGC improves over L2 extend to the right in orange, and pairs where it falls behind extend to the left in blue. Numbers count the pairs in each segment (segments with fewer than $13$ pairs are unlabeled). \textbf{Right:} $\Delta$, the mean per-pair difference in \% (where one seed corresponds to $\frac13$, or $33.3$ points, of a single pair's score), with its 95\% confidence interval.}
\label{fig:pairdiff}
\end{figure}

Table~\ref{tab:success} reports the numerical success rates corresponding to Figure~\ref{fig:success}, and Figure~\ref{fig:pairdiff} breaks down the per-pair differences underlying each $\Delta$ (Appendix~\ref{app:prelim}). In the following subsections, we first benchmark our evaluation pipeline against published results under the original LeWM protocol, and then evaluate NGC across different model checkpoints and training seeds on the pairs of Figure~\ref{fig:success}.

\begin{table}[h]
\centering
\caption{Success rate (\%) of the L2 cost and NGC across the offsets of Figure~\ref{fig:success} (our checkpoint, $300$ pairs per offset, three planner seeds averaged within each pair). Brackets in small type give the 95\% bootstrap confidence interval over pairs (corresponding to the whiskers in Figure~\ref{fig:success}). $\Delta$ is the mean per-pair difference (NGC minus L2) with its 95\% confidence interval, and gained/lost counts the pairs with a positive versus negative difference (Figure~\ref{fig:pairdiff}).}
\label{tab:success}
\small
\setlength{\tabcolsep}{4pt}
\begin{tabular}{lrrrcc}
\toprule
Task & Offset & L2 & NGC & $\Delta$ [95\% CI] & gained/lost \\
\midrule
TwoRoom & 25 & 91.4 {\scriptsize[88.7, 94.0]} & 99.8 {\scriptsize[99.4, 100.0]} & $+8.3$ [$+5.8$, $+11.0$] & 45/1 \\
TwoRoom & 50 & 65.9 {\scriptsize[60.9, 70.9]} & 100.0 & $+34.1$ [$+29.2$, $+39.2$] & 121/0 \\
TwoRoom & 75 & 27.3 {\scriptsize[22.6, 32.1]} & 94.6 {\scriptsize[92.2, 96.7]} & $+67.2$ [$+62.2$, $+72.1$] & 223/0 \\
\midrule
Reacher & 25 & 99.6 {\scriptsize[99.1, 99.9]} & 99.4 {\scriptsize[98.9, 99.9]} & $-0.1$ [$-0.8$, $+0.6$] & 4/5 \\
Reacher & 50 & 95.6 {\scriptsize[93.2, 97.7]} & 100.0 & $+4.4$ [$+2.3$, $+6.9$] & 15/0 \\
Reacher & 100 & 80.6 {\scriptsize[76.1, 84.8]} & 100.0 & $+19.4$ [$+15.2$, $+23.8$] & 68/0 \\
\midrule
Cube & 25 & 47.6 {\scriptsize[42.6, 52.6]} & 57.9 {\scriptsize[53.2, 62.4]} & $+10.3$ [$+6.3$, $+14.4$] & 81/44 \\
Cube & 50 & 42.7 {\scriptsize[37.6, 47.8]} & 53.1 {\scriptsize[48.0, 58.1]} & $+10.4$ [$+6.8$, $+14.2$] & 70/26 \\
Cube & 100 & 34.8 {\scriptsize[30.0, 39.7]} & 50.6 {\scriptsize[45.7, 55.4]} & $+15.8$ [$+11.6$, $+20.0$] & 96/26 \\
\midrule
PushT & 25 & 90.6 {\scriptsize[87.6, 93.3]} & 91.2 {\scriptsize[88.2, 94.0]} & $+0.7$ [$-1.0$, $+2.4$] & 16/13 \\
PushT & 50 & 43.3 {\scriptsize[38.1, 48.6]} & 63.1 {\scriptsize[58.6, 67.7]} & $+19.8$ [$+15.1$, $+24.4$] & 115/32 \\
PushT & 100 & 12.8 {\scriptsize[9.6, 16.2]} & 27.9 {\scriptsize[23.9, 32.0]} & $+15.1$ [$+11.6$, $+18.8$] & 103/19 \\
\bottomrule
\end{tabular}
\end{table}

\subsection{Reproducing the LeWM protocol}
\label{app:comparability}

To compare directly with published numbers, we also replicate the exact random pair-sampling protocol of \citet{maes2026lewm}. This protocol consists of three runs, each drawing $50$ start frames at random from the dataset episodes and taking the frame $25$ steps later in the same episode as the goal ($150$ distinct pairs in total, each evaluated with a single planner seed). Execution and success criteria also follow the original LeWM setup; in particular, each episode initializes the simulator with its recorded nonzero velocity rather than zeroing it out (cf.\ Appendix~\ref{app:protocol}). Our planner configuration differs only in the number of CEM iterations: \citet{maes2026lewm} report $10$ iterations on TwoRoom, Reacher, and Cube and $30$ on PushT, whereas we run $30$ iterations across all tasks, matching their released evaluation code. Episodes selected for evaluation are excluded from graph construction.

Table~\ref{tab:comparability} compares our L2 baseline against the numbers reported by \citet{maes2026lewm}. Using their released checkpoint, our L2 success rates lie $2$ to $7$ points below the published figures: the difference is within one standard deviation across the three runs on TwoRoom, Reacher, and Cube, and slightly larger on PushT ($89.3$\% versus $96$\%). An independent re-evaluation of LeWM by \citet{li2026rcaux} similarly obtains $90.4$\% on PushT and remains within $4$ points of our released-checkpoint L2 numbers across all tasks, showing that a similar PushT discrepancy also occurs in an independent evaluation. Using our own checkpoint, the L2 cost stays within $6$ points of the published numbers on every task.

Crucially, the gains of NGC do not depend on the protocol refinements introduced in Appendix~\ref{app:protocol}. Under the original LeWM protocol, NGC keeps its gain on TwoRoom with both checkpoints and on Cube with our checkpoint, with 95\% intervals excluding zero, and stays within $3$ points of the L2 cost on Reacher and PushT. On Reacher, the L2 cost with our checkpoint drops from $99.6$\% under our protocol (Table~\ref{tab:success}) to $86.0$\% under the LeWM protocol. Two features of the LeWM protocol can contribute to failures: the initial velocity is not observable from the start image, and comparing raw shoulder angles treats identical poses separated by a $2\pi$ rotation as failures. Because the pair sets and other protocol settings also differ, this comparison does not isolate the contribution of either feature. On Cube with the released checkpoint, NGC gains less under the LeWM protocol ($+6.0$) than at offset $25$ of our protocol ($+11.2$, Table~\ref{tab:ckpt}) because random sampling retains pairs that already satisfy the success criterion at the start state ($38$\% of sampled Cube pairs; \citealp{sun2026intact}), which trivially succeed under both costs.

\begin{table}[h]
\centering
\caption{Evaluation under the original LeWM protocol. Success (\%) is reported as the mean $\pm$ SD across three runs of $50$ random pairs each. The last column lists the integer-rounded figures reported by \citet{maes2026lewm} (their Figure~6). $\Delta$ is the mean per-pair difference across all $150$ pairs (NGC minus L2) with its 95\% confidence interval (Appendix~\ref{app:prelim}).}
\label{tab:comparability}
\footnotesize
\setlength{\tabcolsep}{2pt}
\begin{tabular}{lccccccc}
\toprule
& \multicolumn{3}{c}{our checkpoint} & \multicolumn{3}{c}{released LeWM checkpoint} & \\
\cmidrule(lr){2-4}\cmidrule(lr){5-7}
Task & L2 & NGC & $\Delta$ {\scriptsize[95\% CI]} & L2 & NGC & $\Delta$ {\scriptsize[95\% CI]} & reported \\
\midrule
TwoRoom & $91.3$ {\scriptsize$\pm3.1$} & $100.0$ {\scriptsize$\pm0.0$} & $+8.7$ {\scriptsize$[+4.7,+13.3]$} & $85.3$ {\scriptsize$\pm3.1$} & $100.0$ {\scriptsize$\pm0.0$} & $+14.7$ {\scriptsize$[+9.3,+20.7]$} & $87$ \\
Reacher & $86.0$ {\scriptsize$\pm2.0$} & $84.0$ {\scriptsize$\pm3.5$} & $-2.0$ {\scriptsize$[-9.3,+5.3]$} & $82.7$ {\scriptsize$\pm4.2$} & $81.3$ {\scriptsize$\pm11.4$} & $-1.3$ {\scriptsize$[-9.3,+6.7]$} & $86$ \\
Cube & $68.0$ {\scriptsize$\pm10.6$} & $78.7$ {\scriptsize$\pm7.6$} & $+10.7$ {\scriptsize$[+4.7,+17.3]$} & $69.3$ {\scriptsize$\pm12.1$} & $75.3$ {\scriptsize$\pm5.0$} & $+6.0$ {\scriptsize$[-0.7,+12.7]$} & $74$ \\
PushT & $91.3$ {\scriptsize$\pm2.3$} & $94.0$ {\scriptsize$\pm5.3$} & $+2.7$ {\scriptsize$[0.0,+6.0]$} & $89.3$ {\scriptsize$\pm5.0$} & $88.7$ {\scriptsize$\pm8.1$} & $-0.7$ {\scriptsize$[-4.7,+3.3]$} & $96$ \\
\bottomrule
\end{tabular}
\end{table}

\subsection{The released LeWM checkpoint}
\label{app:ckpt}

Table~\ref{tab:ckpt} repeats the evaluation of Figure~\ref{fig:success} using the official checkpoint released by \citet{maes2026lewm} in place of ours. The evaluation pairs, planner seeds, success criteria, and step limits remain identical, with each pair seed's graph encoded by the released checkpoint. The pattern of improvements carries over: except at offset $25$ on Reacher and PushT, NGC outperforms the L2 cost on every task and offset, with 95\% intervals excluding zero. At those two settings, the rates differ by less than $1$ point and the intervals include zero. Across nearly all settings, the gains match those of our checkpoint within $4$ points; the only exceptions are TwoRoom at offset $25$ (where the released checkpoint starts from a lower L2 success rate and thus gains more) and PushT at offset $100$ (where it gains $+8.1$ versus $+15.1$).

\subsection{Training seeds}
Table~\ref{tab:seeds} repeats the evaluation of Figure~\ref{fig:success} across two additional training seeds of our training recipe. The pairs, planner seeds, success criteria, and step limits are unchanged, and each pair seed's graph is encoded by the corresponding checkpoint. Both additional training seeds closely reproduce the gains of our primary checkpoint: across all tasks and offsets, $\Delta$ stays within $4$ points of the first seed, except at offset $75$ on TwoRoom, where the third seed gains $+55.0$ compared with $+67.2$. There, one of the three graphs of the third seed contains two edges that cross the wall (Appendix~\ref{app:graphsens}), and NGC solves $69$\% of that pair set, against $93$ to $98$\% of the other eight. Within that affected pair set, NGC solves only $16$\% of the $33$ pairs whose shortest graph path routes through the wall-crossing edges, versus $96$\% of the remaining $67$ pairs.

\begin{table}[h]
\centering
\caption{Performance of the released LeWM checkpoint on the pairs of Figure~\ref{fig:success} ($300$ pairs per offset, three planner seeds averaged within each pair). $\Delta$ is the mean per-pair difference (NGC minus L2) with its 95\% confidence interval; the final column repeats $\Delta$ from Table~\ref{tab:success} for our checkpoint.}
\label{tab:ckpt}
\small
\begin{tabular}{llrrcr}
\toprule
Task & Offset & L2 & NGC & $\Delta$ [95\% CI] & $\Delta$, ours \\
\midrule
TwoRoom & 25 & 85.3 & 99.8 & $+14.4$ [$+11.6$, $+17.6$] & $+8.3$ \\
TwoRoom & 50 & 61.9 & 99.8 & $+37.9$ [$+32.9$, $+42.9$] & $+34.1$ \\
TwoRoom & 75 & 28.0 & 95.1 & $+67.1$ [$+62.2$, $+72.0$] & $+67.2$ \\
\midrule
Reacher & 25 & 97.9 & 97.4 & $-0.4$ [$-1.8$, $+1.0$] & $-0.1$ \\
Reacher & 50 & 96.2 & 100.0 & $+3.8$ [$+1.9$, $+5.9$] & $+4.4$ \\
Reacher & 100 & 82.6 & 100.0 & $+17.4$ [$+13.4$, $+21.6$] & $+19.4$ \\
\midrule
Cube & 25 & 44.7 & 55.9 & $+11.2$ [$+7.4$, $+15.1$] & $+10.3$ \\
Cube & 50 & 43.4 & 57.9 & $+14.4$ [$+10.4$, $+18.6$] & $+10.4$ \\
Cube & 100 & 35.0 & 51.1 & $+16.1$ [$+12.1$, $+20.2$] & $+15.8$ \\
\midrule
PushT & 25 & 84.7 & 83.9 & $-0.8$ [$-2.8$, $+1.1$] & $+0.7$ \\
PushT & 50 & 40.0 & 57.8 & $+17.8$ [$+13.7$, $+22.1$] & $+19.8$ \\
PushT & 100 & 12.2 & 20.3 & $+8.1$ [$+5.0$, $+11.3$] & $+15.1$ \\
\bottomrule
\end{tabular}
\end{table}

\begin{table}[h]
\centering
\caption{Evaluation across two additional world model training seeds on the pairs of Figure~\ref{fig:success} ($300$ pairs per offset, three planner seeds averaged within each pair). Columns report success rates (\%) of the L2 cost and NGC under each seed, along with the mean per-pair difference $\Delta$ (NGC minus L2) and its 95\% confidence interval. The final column repeats $\Delta$ from Table~\ref{tab:success} for our primary checkpoint (first seed).}
\label{tab:seeds}
\small
\setlength{\tabcolsep}{4pt}
\begin{tabular}{lrrrcrrcr}
\toprule
 & & \multicolumn{3}{c}{Second seed} & \multicolumn{3}{c}{Third seed} & First seed \\
\cmidrule(lr){3-5}\cmidrule(lr){6-8}\cmidrule(lr){9-9}
Task & Offset & L2 & NGC & $\Delta$ [95\% CI] & L2 & NGC & $\Delta$ [95\% CI] & $\Delta$ \\
\midrule
TwoRoom & 25 & 92.2 & 100.0 & $+7.8$ {\scriptsize$[+5.4,+10.3]$} & 92.6 & 100.0 & $+7.4$ {\scriptsize$[+5.1,+10.0]$} & $+8.3$ \\
TwoRoom & 50 & 69.7 & 99.8 & $+30.1$ {\scriptsize$[+25.4,+34.9]$} & 66.3 & 99.2 & $+32.9$ {\scriptsize$[+27.9,+38.1]$} & $+34.1$ \\
TwoRoom & 75 & 31.8 & 96.1 & $+64.3$ {\scriptsize$[+59.3,+69.2]$} & 31.4 & 86.4 & $+55.0$ {\scriptsize$[+49.7,+60.3]$} & $+67.2$ \\
\midrule
Reacher & 25 & 99.4 & 99.4 & $0.0$ {\scriptsize$[-0.7,+0.7]$} & 99.7 & 99.9 & $+0.2$ {\scriptsize$[-0.2,+0.7]$} & $-0.1$ \\
Reacher & 50 & 95.4 & 100.0 & $+4.6$ {\scriptsize$[+2.4,+7.0]$} & 95.6 & 99.9 & $+4.3$ {\scriptsize$[+2.2,+6.7]$} & $+4.4$ \\
Reacher & 100 & 80.7 & 100.0 & $+19.3$ {\scriptsize$[+15.1,+23.7]$} & 81.6 & 100.0 & $+18.4$ {\scriptsize$[+14.3,+22.7]$} & $+19.4$ \\
\midrule
Cube & 25 & 48.8 & 61.3 & $+12.6$ {\scriptsize$[+8.6,+16.7]$} & 46.3 & 57.9 & $+11.6$ {\scriptsize$[+7.9,+15.3]$} & $+10.3$ \\
Cube & 50 & 43.7 & 56.4 & $+12.8$ {\scriptsize$[+8.9,+16.8]$} & 44.6 & 53.6 & $+9.0$ {\scriptsize$[+5.4,+12.7]$} & $+10.4$ \\
Cube & 100 & 36.2 & 53.7 & $+17.4$ {\scriptsize$[+13.0,+22.0]$} & 34.7 & 51.3 & $+16.7$ {\scriptsize$[+12.7,+20.8]$} & $+15.8$ \\
\midrule
PushT & 25 & 90.4 & 91.3 & $+0.9$ {\scriptsize$[-1.0,+2.9]$} & 88.8 & 89.1 & $+0.3$ {\scriptsize$[-1.8,+2.6]$} & $+0.7$ \\
PushT & 50 & 41.6 & 59.9 & $+18.3$ {\scriptsize$[+13.7,+23.0]$} & 43.1 & 61.2 & $+18.1$ {\scriptsize$[+13.4,+22.9]$} & $+19.8$ \\
PushT & 100 & 12.6 & 24.9 & $+12.3$ {\scriptsize$[+8.9,+15.9]$} & 13.1 & 24.4 & $+11.3$ {\scriptsize$[+7.9,+14.8]$} & $+15.1$ \\
\bottomrule
\end{tabular}
\end{table}

\FloatBarrier

\section{Sensitivity to the graph construction}
\label{app:graphsens}

\paragraph{Node sample and neighbor rule.} Table~\ref{tab:graphsens} varies the default graph construction on the first pair seed of Table~\ref{tab:success} along two axes: node sampling and the neighbor connectivity rule. The two resamples each draw $N=5{,}000$ new random frames from the same pool of eligible episodes while keeping the mutual-15NN rule, whereas the \emph{union} variant keeps the default nodes but connects any pair of nodes where either belongs to the other's $15$ nearest neighbors. Across all tasks and offsets, every 95\% interval of $\Delta$ against the default graph includes zero for both node resamples and the union rule, while at the longer offset of each task every variant retains a higher success rate than the L2 cost. Offline, replacing the mutual rule with the union rule leaves the Spearman correlation between NGC and physical distance within $0.02$ of the default ($0.94$ on TwoRoom, $0.99$ on Reacher, $0.73$ versus $0.75$ on Cube, and $0.54$ versus $0.55$ on PushT), but introduces longer edges: on TwoRoom, the longest edge spans $47$\,px of physical path length under the union rule, compared with $29$\,px under the mutual rule. We use the mutual rule as our default because it enforces stricter locality, aligning with the manifold hypothesis.

\begin{table}[h]
\centering
\caption{Sensitivity to node sampling and the neighbor connectivity rule on one pair seed ($100$ pairs $\times$ $3$ planner seeds per cell). All columns except L2 evaluate NGC. The L2 and Default columns (mutual rule, $k=15$, $5{,}000$ nodes) correspond to the first pair seed of Table~\ref{tab:success}. Resamples~1 and~2 draw new sets of $5{,}000$ nodes under the mutual rule; the union column uses the default nodes but joins two nodes when either is among the other's $15$ nearest neighbors. Each variant cell reports the success rate (\%) and, in small type, the mean per-pair difference $\Delta$ against the default graph with its 95\% confidence interval.}
\label{tab:graphsens}
\footnotesize
\begin{tabular}{llccccc}
\toprule
& & & \multicolumn{3}{c}{Node sample (mutual)} & Neighbor rule \\
\cmidrule(lr){4-6}\cmidrule(lr){7-7}
Task & Offset & L2 & Default & resample 1 & resample 2 & union \\
\midrule
TwoRoom & 25 & 91.3 & 100.0 & \shortstack{100.0\\{\scriptsize$0.0$}} & \shortstack{100.0\\{\scriptsize$0.0$}} & \shortstack{100.0\\{\scriptsize$0.0$}} \\
TwoRoom & 75 & 23.7 & 95.7 & \shortstack{95.3\\{\scriptsize$-0.3\,[-3.0,+2.3]$}} & \shortstack{94.7\\{\scriptsize$-1.0\,[-3.7,+1.3]$}} & \shortstack{94.3\\{\scriptsize$-1.3\,[-4.0,+1.0]$}} \\
\midrule
Reacher & 25 & 99.0 & 99.0 & \shortstack{99.3\\{\scriptsize$+0.3\,[-1.0,+1.7]$}} & \shortstack{98.7\\{\scriptsize$-0.3\,[-2.0,+1.3]$}} & \shortstack{100.0\\{\scriptsize$+1.0\,[0.0,+2.3]$}} \\
Reacher & 100 & 82.7 & 100.0 & \shortstack{100.0\\{\scriptsize$0.0$}} & \shortstack{100.0\\{\scriptsize$0.0$}} & \shortstack{100.0\\{\scriptsize$0.0$}} \\
\midrule
Cube & 25 & 43.7 & 57.0 & \shortstack{54.3\\{\scriptsize$-2.7\,[-8.3,+2.7]$}} & \shortstack{53.0\\{\scriptsize$-4.0\,[-9.7,+1.3]$}} & \shortstack{56.3\\{\scriptsize$-0.7\,[-5.3,+4.0]$}} \\
Cube & 100 & 35.3 & 49.7 & \shortstack{51.0\\{\scriptsize$+1.3\,[-3.3,+6.0]$}} & \shortstack{53.0\\{\scriptsize$+3.3\,[-2.0,+8.7]$}} & \shortstack{50.3\\{\scriptsize$+0.7\,[-3.3,+4.7]$}} \\
\midrule
PushT & 25 & 88.7 & 89.0 & \shortstack{90.7\\{\scriptsize$+1.7\,[-1.3,+5.0]$}} & \shortstack{88.7\\{\scriptsize$-0.3\,[-3.3,+2.7]$}} & \shortstack{88.7\\{\scriptsize$-0.3\,[-2.0,+1.3]$}} \\
PushT & 100 & 13.0 & 28.3 & \shortstack{33.0\\{\scriptsize$+4.7\,[-2.3,+11.7]$}} & \shortstack{33.0\\{\scriptsize$+4.7\,[-2.3,+12.0]$}} & \shortstack{26.7\\{\scriptsize$-1.7\,[-6.3,+3.0]$}} \\
\bottomrule
\end{tabular}
\end{table}

\paragraph{Graph size and neighborhood size.} Table~\ref{tab:graphk} varies the total number of graph nodes $N$ and the neighborhood size $k$ under the mutual rule. On Reacher, Cube, and PushT, NGC is largely insensitive to both hyperparameters: except when $N$ is reduced to $\frac18$ of the default ($625$ nodes), every 95\% interval of $\Delta$ against the default graph includes zero and lies within $-11.3$ to $+8.3$ points. On TwoRoom, however, the variants diverge sharply at offset $75$: smaller neighborhoods ($k=5$ and $10$) match the high success rate of the default graph, whereas sparser node sets or larger neighborhoods ($k=30$ and $60$) drop by $22$ to $51$ points.

\begin{figure}[h]
\centering
\includegraphics[width=\linewidth]{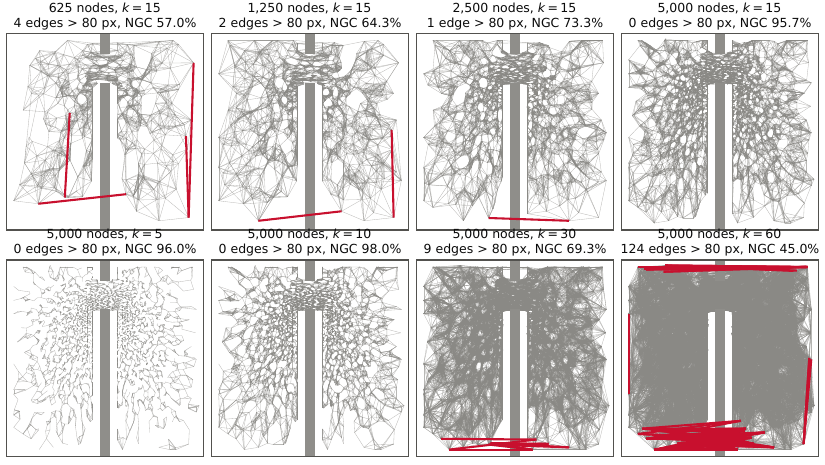}
\caption{Physical-space visualizations of the TwoRoom graphs from Table~\ref{tab:graphk}: nested node subsets and the default graph at $k=15$ (\textbf{top}), and the default $5{,}000$ nodes at $k=5$, $10$, $30$, and $60$ (\textbf{bottom}), with each edge plotted between the physical positions of its endpoint nodes. Red edges highlight shortcuts that connect nodes separated by more than $80$\,px along the shortest feasible path through the doorway. Titles report the number of such shortcut edges and the resulting NGC success rate at offset $75$.}
\label{fig:graphsize}
\end{figure}

Figure~\ref{fig:graphsize} illustrates a possible geometric explanation for this drop by plotting each TwoRoom graph from Table~\ref{tab:graphk} in physical space, with edges drawn between the rendered $(x,y)$ positions of their endpoint nodes. The five variants whose performance degrades at offset $75$ are precisely the five whose graphs contain shortcut edges across the central wall, linking nodes that happen to lie relatively close in the ambient latent space despite being far apart along the feasible path through the doorway; the default graph ($N=5{,}000, k=15$) and the smaller-$k$ variants ($k=5, 10$) contain no such edges. At $\frac12 N$ ($2{,}500$ nodes), the graph contains a single wall-crossing edge and the success rate is $22$ points below the default. This association is consistent with harmful shortcuts, but changing the node set also changes coverage, attachment, and other edges, so the comparison does not isolate the effect of that edge. Because the mutual-$k$NN rule connects two nodes only when each falls within the other's $k$-nearest neighborhood, the spatial reach of an edge scales with the distance to the $k$-th nearest neighbor. Either reducing the node density $N$ or increasing $k$ expands this reach until it bridges the Euclidean gap across the wall in latent space, causing shortest paths on the graph to cut through the wall.

\begin{table}[h]
\centering
\caption{Sensitivity to graph size $N$ and neighborhood size $k$ on one pair seed ($100$ pairs $\times$ $3$ planner seeds per cell). Column headings indicate goal offsets in steps. Each cell reports the success rate (\%) and, in small type, the mean per-pair difference $\Delta$ against the default graph ($5{,}000$ nodes, $k=15$) with its 95\% confidence interval.}
\label{tab:graphk}
\footnotesize
\setlength{\tabcolsep}{3pt}
\begin{tabular}{lcccc}
\toprule
& \multicolumn{2}{c}{TwoRoom} & \multicolumn{2}{c}{Reacher} \\
\cmidrule(lr){2-3}\cmidrule(lr){4-5}
Variant & 25 & 75 & 25 & 100 \\
\midrule
L2 & 91.3 & 23.7 & 99.0 & 82.7 \\
NGC (default) & 100.0 & 95.7 & 99.0 & 100.0 \\
\midrule
$N=625$ & \shortstack{100.0\\{\scriptsize$0.0$}} & \shortstack{57.0\\{\scriptsize$-38.7\,[-48.0,-29.7]$}} & \shortstack{99.7\\{\scriptsize$+0.7\,[-0.7,+2.0]$}} & \shortstack{100.0\\{\scriptsize$0.0$}} \\
$N=1{,}250$ & \shortstack{100.0\\{\scriptsize$0.0$}} & \shortstack{64.3\\{\scriptsize$-31.3\,[-40.0,-23.0]$}} & \shortstack{99.7\\{\scriptsize$+0.7\,[-0.7,+2.0]$}} & \shortstack{100.0\\{\scriptsize$0.0$}} \\
$N=2{,}500$ & \shortstack{100.0\\{\scriptsize$0.0$}} & \shortstack{73.3\\{\scriptsize$-22.3\,[-30.3,-14.7]$}} & \shortstack{99.7\\{\scriptsize$+0.7\,[0.0,+1.7]$}} & \shortstack{100.0\\{\scriptsize$0.0$}} \\
\midrule
$k=5$ & \shortstack{100.0\\{\scriptsize$0.0$}} & \shortstack{96.0\\{\scriptsize$+0.3\,[-1.0,+2.0]$}} & \shortstack{99.3\\{\scriptsize$+0.3\,[-1.0,+1.7]$}} & \shortstack{100.0\\{\scriptsize$0.0$}} \\
$k=10$ & \shortstack{99.7\\{\scriptsize$-0.3\,[-1.0,0.0]$}} & \shortstack{98.0\\{\scriptsize$+2.3\,[+0.3,+5.0]$}} & \shortstack{99.0\\{\scriptsize$0.0\,[-1.7,+1.7]$}} & \shortstack{100.0\\{\scriptsize$0.0$}} \\
$k=30$ & \shortstack{100.0\\{\scriptsize$0.0$}} & \shortstack{69.3\\{\scriptsize$-26.3\,[-34.7,-18.3]$}} & \shortstack{100.0\\{\scriptsize$+1.0\,[0.0,+2.3]$}} & \shortstack{100.0\\{\scriptsize$0.0$}} \\
$k=60$ & \shortstack{95.3\\{\scriptsize$-4.7\,[-9.0,-1.0]$}} & \shortstack{45.0\\{\scriptsize$-50.7\,[-60.0,-41.3]$}} & \shortstack{99.7\\{\scriptsize$+0.7\,[-0.7,+2.0]$}} & \shortstack{100.0\\{\scriptsize$0.0$}} \\
\midrule
& \multicolumn{2}{c}{Cube} & \multicolumn{2}{c}{PushT} \\
\cmidrule(lr){2-3}\cmidrule(lr){4-5}
Variant & 25 & 100 & 25 & 100 \\
\midrule
L2 & 43.7 & 35.3 & 88.7 & 13.0 \\
NGC (default) & 57.0 & 49.7 & 89.0 & 28.3 \\
\midrule
$N=625$ & \shortstack{50.3\\{\scriptsize$-6.7\,[-13.0,-0.7]$}} & \shortstack{46.3\\{\scriptsize$-3.3\,[-9.0,+2.0]$}} & \shortstack{88.7\\{\scriptsize$-0.3\,[-3.0,+2.0]$}} & \shortstack{20.3\\{\scriptsize$-8.0\,[-14.3,-2.0]$}} \\
$N=1{,}250$ & \shortstack{56.3\\{\scriptsize$-0.7\,[-6.0,+5.0]$}} & \shortstack{47.3\\{\scriptsize$-2.3\,[-7.0,+2.3]$}} & \shortstack{86.7\\{\scriptsize$-2.3\,[-5.7,+0.7]$}} & \shortstack{22.7\\{\scriptsize$-5.7\,[-11.3,+0.3]$}} \\
$N=2{,}500$ & \shortstack{55.7\\{\scriptsize$-1.3\,[-6.7,+4.0]$}} & \shortstack{53.0\\{\scriptsize$+3.3\,[-1.3,+8.0]$}} & \shortstack{90.0\\{\scriptsize$+1.0\,[-1.7,+4.3]$}} & \shortstack{27.0\\{\scriptsize$-1.3\,[-8.0,+5.7]$}} \\
\midrule
$k=5$ & \shortstack{56.7\\{\scriptsize$-0.3\,[-5.3,+4.7]$}} & \shortstack{51.0\\{\scriptsize$+1.3\,[-3.0,+5.7]$}} & \shortstack{89.0\\{\scriptsize$0.0\,[-4.3,+4.3]$}} & \shortstack{29.7\\{\scriptsize$+1.3\,[-5.7,+8.3]$}} \\
$k=10$ & \shortstack{58.0\\{\scriptsize$+1.0\,[-4.0,+6.0]$}} & \shortstack{49.0\\{\scriptsize$-0.7\,[-5.3,+4.3]$}} & \shortstack{90.3\\{\scriptsize$+1.3\,[-1.3,+4.7]$}} & \shortstack{29.7\\{\scriptsize$+1.3\,[-4.7,+7.3]$}} \\
$k=30$ & \shortstack{52.7\\{\scriptsize$-4.3\,[-9.7,+1.0]$}} & \shortstack{50.3\\{\scriptsize$+0.7\,[-2.7,+4.3]$}} & \shortstack{89.0\\{\scriptsize$0.0\,[-2.0,+2.0]$}} & \shortstack{26.3\\{\scriptsize$-2.0\,[-8.7,+4.7]$}} \\
$k=60$ & \shortstack{52.7\\{\scriptsize$-4.3\,[-10.7,+1.7]$}} & \shortstack{48.7\\{\scriptsize$-1.0\,[-6.0,+3.7]$}} & \shortstack{88.7\\{\scriptsize$-0.3\,[-2.7,+2.0]$}} & \shortstack{26.3\\{\scriptsize$-2.0\,[-8.0,+4.0]$}} \\
\bottomrule
\end{tabular}
\end{table}

\begin{figure}[t]
\centering
\includegraphics[width=\linewidth]{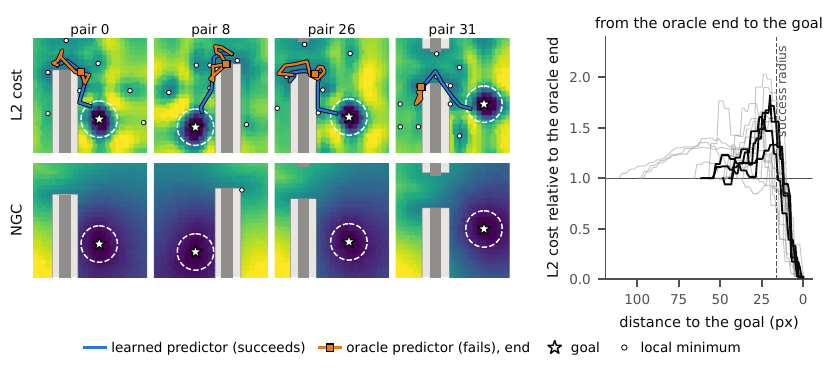}
\caption{Behavior of the L2 cost with the oracle predictor on TwoRoom at offset $25$. \textbf{Left:} the first four pairs (in pair-index order) where the L2 cost fails under the oracle predictor but succeeds under the learned predictor, cropped around the stopping point and the goal. Top row: L2 cost maps overlaid with rollouts from both predictors. Bottom row: corresponding NGC cost maps. Each panel is normalized to its own 2nd--98th percentile within the crop; the dashed circle marks the goal success radius, and white dots indicate strict local minima on the grid (points cheaper than all $8$ neighbors). \textbf{Right:} L2 cost profiles along the straight line from the oracle rollout's stopping point to the goal, normalized by the cost at the stopping point, across all $16$ pairs where the L2 cost fails with the oracle predictor (the four pairs shown on the left are highlighted in black).}
\label{fig:oraclel2}
\end{figure}

\section{Why can an oracle predictor lower success on TwoRoom?}
\label{app:oraclel2}

In Table~\ref{tab:oracle}, replacing the learned predictor with the oracle predictor slightly reduces the success rate of the L2 cost at every offset on TwoRoom. To understand why exact predictions can hurt planning under the L2 cost, we inspect the rollouts and cost maps at offset $25$.

\paragraph{Cost geometry near the oracle stopping points.} Because the oracle predictor scores each candidate sequence by rendering and encoding its exact terminal state, the cost it assigns matches the ground-truth L2 cost map from Figure~\ref{fig:costmaps} (up to the $4$\,px grid stride). Figure~\ref{fig:oraclel2} displays the cost maps and rollouts for four pairs where the L2 cost fails with the oracle predictor but succeeds with the learned predictor (left), alongside one-dimensional cost profiles for all $16$ oracle failures (right). Moving along the straight path from each oracle stopping point toward the goal, the true L2 cost initially rises above its value at the stopping point, dropping below it only within the immediate vicinity of the goal. Consequently, advancing toward the goal from these positions actually worsens the true L2 objective before improving it.

\paragraph{Why exact predictions need not improve success.} Of the $16$ pairs where the oracle predictor fails under the L2 cost, the learned predictor solves $11$. Across all $11$ pairs, the oracle rollout stops beside the doorway, $32$ to $68$\,px from the goal (median $50$\,px), and never approaches closer than $26$\,px. A possible explanation is that prediction errors make a candidate that advances toward the goal appear cheaper than one that stays near the stopping point, even when their true terminal L2 costs favor staying. Removing such a favorable ranking error can reduce success despite making predictions exact.


\end{document}